\documentclass[10pt,twocolumn,letterpaper]{article}

\usepackage[pagenumbers]{wacv} 

\usepackage{graphicx}
\usepackage{amsmath}
\usepackage{amssymb}
\usepackage{booktabs}
\usepackage[usenames,dvipsnames]{xcolor}

\newcommand{\algname}{\textsc{SyncDiff }}

\newcommand{\first}[1]{{\textbf{\color{black}#1}}}
\newcommand{\second}[1]{{\textbf{\color{violet}#1}}}

\usepackage[pagebackref,breaklinks,colorlinks]{hyperref}

\usepackage[capitalize]{cleveref}
\crefname{section}{Sec.}{Secs.}
\Crefname{section}{Section}{Sections}
\Crefname{table}{Table}{Tables}
\crefname{table}{Tab.}{Tabs.}

\def\wacvPaperID{2251} 
\def\confName{WACV}
\def\confYear{2025}

\begin{document}

\title{\textsc{SyncDiff}: Diffusion-based Talking Head Synthesis with Bottlenecked Temporal Visual Prior for Improved Synchronization}

\author{
    Xulin Fan$^{1}\thanks{Equal contribution.}$\hfill Heting Gao$^{1*}$\hfill Ziyi Chen$^{2}$\hfill Peng Chang$^{2}$\hfill Mei Han$^{2}$\hfill Mark Hasegawa-Johnson$^{1}$\\
    $^{1}$University of Illinois at Urbana-Champaign \hspace{15pt} $^{2}$PAII Inc.\\
    \tt\small{\{xulinf2, hgao17, jhasegaw\}@illinois.edu \hfill\{chenziyi253, changpeng805, hanmei613\}@paii-labs.com}%
}

\maketitle

\begin{abstract}
Talking head synthesis, also known as speech-to-lip synthesis, reconstructs the facial motions that align with the given audio tracks.
The synthesized videos are evaluated on mainly two aspects, lip-speech synchronization and image fidelity. 
Recent studies demonstrate that GAN-based and diffusion-based models achieve state-of-the-art (SOTA) performance on this task, with diffusion-based models achieving superior image fidelity but experiencing lower synchronization compared to their GAN-based counterparts.
To this end, we propose \textsc{SyncDiff}, a simple yet effective approach to improve diffusion-based models 
using a temporal pose frame with information bottleneck and facial-informative audio features extracted from \textsc{AVHuBERT}, as conditioning input into the diffusion process.
We evaluate \textsc{SyncDiff} on two canonical talking head datasets, LRS2 and LRS3 for direct comparison with other SOTA models. 
Experiments on LRS2$/$LRS3 datasets show that \textsc{SyncDiff} achieves a synchronization score $27.7\%/62.3\%$ relatively higher than previous diffusion-based methods, while preserving their high-fidelity characteristics.
\end{abstract}

\section{Introduction}
\label{sec:intro}
Talking head synthesis has gained popularity as a research topic due to its expanding applications, which include virtual being creation, online conferences, audio dubbing, and video translation. Its primary objective is to generate lip-synced avatar videos based on given speech audio~\cite{kim2018deep,chen2019hierarchical,jamaludin2019you,zhou2020makelttalk,chen2020talking,wang2021one}.

Over recent years, many studies have been conducted to synthesize realistic talking head videos. Some approaches divide the entire generation process into two steps, utilizing facial landmarks or facial-model-based parameters as an intermediate feature. The first step is designed to predict precise intermediate features from speech audio, and the second step focuses on generating realistic images given the intermediate features~\cite{chen2019hierarchical,zhou2020makelttalk, chen2020talking,lu2021live,Lahiri_2021_CVPR,ye2023geneface}. Recently, more works have focused on designing end-to-end video generation with advanced generative models~\cite{KR2020ALS,zhou2019talking,guo2021adnerf,Wang2023SeeingWY,Du2023DAETalkerHF,Zhang2022SadTalkerLR,Shen2023DiffTalkCD,Stylesync_cvpr23}.
In addition to variations in the talking head generation pipeline, studies also vary in model generality. Some approaches are character-specific, necessitating the model to undergo training or adaptation using a substantial amount of data specific to the particular character~\cite{lu2021live,guo2021adnerf,Du2023DAETalkerHF,ye2023geneface}. On the other hand, some models adopt a more general synthesis approach without character restrictions~\cite{zhou2019talking,KR2020ALS,Wang2023SeeingWY,Zhang2022SadTalkerLR,Shen2023DiffTalkCD,Stylesync_cvpr23}. These models are trained with large talking head datasets across various characters. This approach grants them the zero-shot ability to synthesize new character videos without additional character-specific training data and retraining the model.

A well-synchronized lip movement and high-fidelity generated image are the two most critical aspects of talking head synthesis. Several Generative Adversarial Network (GAN) based studies featuring a specific lip synchronization training design have demonstrated notable results in voice-lip synchronizations~\cite{KR2020ALS,Wang2023SeeingWY,Stylesync_cvpr23}. Nevertheless, these GAN-based approaches are beset by challenges such as unstable training processes and suboptimal visual quality. Conversely, certain diffusion-based models ~\cite{Du2023DAETalkerHF,Shen2023DiffTalkCD} exhibit superior image quality, yet their generated videos may not attain the same degree of lip synchronization as observed in GAN-based models. More recently, diffusion-based models~\cite{tian2024emo, xu2024hallo} focused on whole face generation were proposed with better image quality and temporal coherence. However, both models employ complex motion control modules and require large-scale internet-sourced data, which significantly hinders training and inference speed.

To address the limitations inherent in the aforementioned directions, we introduce \textsc{SyncDiff}, a diffusion model that incorporates temporal-augmented visual priors to condition the diffusion process~\cite{rombach2022high}, focused on lip region inpainting without the need for crowd-sourced data. Additionally, our model integrates a pre-trained audio-video self-supervised framework to enhance the synchronization of the generated videos.

Many studies leverage temporal information by employing a temporally aware audio encoder, yet they solely utilize a single image for identity information~\cite{Wang2023SeeingWY,Zhang2022SadTalkerLR,Shen2023DiffTalkCD,Du2023DAETalkerHF,ye2023geneface}, thereby neglecting the temporal information present in the video modality.  Due to the high similarity within short-term temporal images, directly incorporating neighborhood image information may lead to training shortcuts~\cite{KR2020ALS,Shen2023DiffTalkCD}. In response to this concern, we propose a bottleneck layer to leverage temporal pose information and mitigate the risk of training shortcuts.

We conduct comprehensive experiments on two benchmark datasets, demonstrating that our proposed \textsc{SyncDiff} attains superior visual quality compared to all state-of-the-art synthesis methods. Additionally, our approach significantly improves lip synchronization within the diffusion-based synthesis model, showing competitive performance compared to other advanced lip synchronization methods. In summary, our primary contributions are as follows:
\begin{itemize}
  \item We present a novel approach by introducing a bottleneck layer to incorporate visual temporal pose information in talking head synthesis. This pioneering addition of visual temporal pose information results in a substantial improvement in lip synchronization.
  \item We leverage a self-supervised audio-visual pre-trained model, \textsc{AVHuBERT}~\cite{shi2022avhubert,shi2022avsr}, within the diffusion model to facilitate talking head generation. This integration contributes to further enhancing lip synchronization.
  \item Our proposed \textsc{SyncDiff} conditional diffusion model excels in generating high-fidelity image quality and well-synchronized lip movement. Our experimental results demonstrate that \textsc{SyncDiff} surpasses other SOTA methods in terms of visual quality while significantly enhancing lip synchronization capabilities in diffusion models.

\end{itemize}
\section{Related Work}
\label{sec:related_work}

\subsection{2D Audio-driven Talking Head Synthesis}

Previous works on talking head synthesis can be categorized into two categories, 3D-based methods, which model in the 3D facial geometry space, and 2D based methods, which directly operate in the 2D pixel space, irrespective of the underlying 3D geometry.

The most relevant previous works are 2D-based methods with generative adversarial networks (GANs) emerging as a favored choice for talking head synthesis focused on lip region impainting~\cite{KR2020ALS,Wang2023SeeingWY, Stylesync_cvpr23}. \textsc{Wav2Lip}~\cite{KR2020ALS} proposes to leverage a cross-modality sync loss from a pretrained lip-sync expert to improve synchronization, whose synchronization performance is still among the highest nowadays. \textsc{TalkLip}~\cite{Wang2023SeeingWY}, one of the latest GAN-based methods, proposes to further incorporate lipreading loss which explicitly aligns the synthetic video with the text modality and further improves the video's lip-reading intelligibility, which is measured by the word error rate of lipreading the synthetic videos. \textsc{StyleSync}~\cite{Stylesync_cvpr23} leverages a \textsc{StyleGAN}~\cite{styleGan2_cvpr2020} based architecture to solve the task by encoding the facial and audio information into the style space. In addition to GAN-based methods, diffusion-based methods emerge as an alternative approach to the task~\cite{Shen2023DiffTalkCD, Du2023DAETalkerHF}. \textsc{DAETalker}~\cite{Du2023DAETalkerHF} trains a speech-to-latent unit whose target is the visual embedding from the video stream. During inference, the latent from input speech is fed into a diffusion decoder to synthesize the video. However, this approach needs to train an identity-specific speech-to-latent encoder and does not generalize to unseen identities. \textsc{DiffTalk}~\cite{Shen2023DiffTalkCD} is another work that proposes to use a combination of a masked frame and a reference frame to condition a diffusion model to synthesize the output. This approach generates high-fidelity videos and generalizes to unseen identities with a given reference image. However, it lags behind the GAN-based models in terms of synchronization, as measured by the \textsc{SyncNet}~\cite{Chung16a} score and lipreading word error rate. 
\begin{figure*}[ht]
          \includegraphics[width=\textwidth]{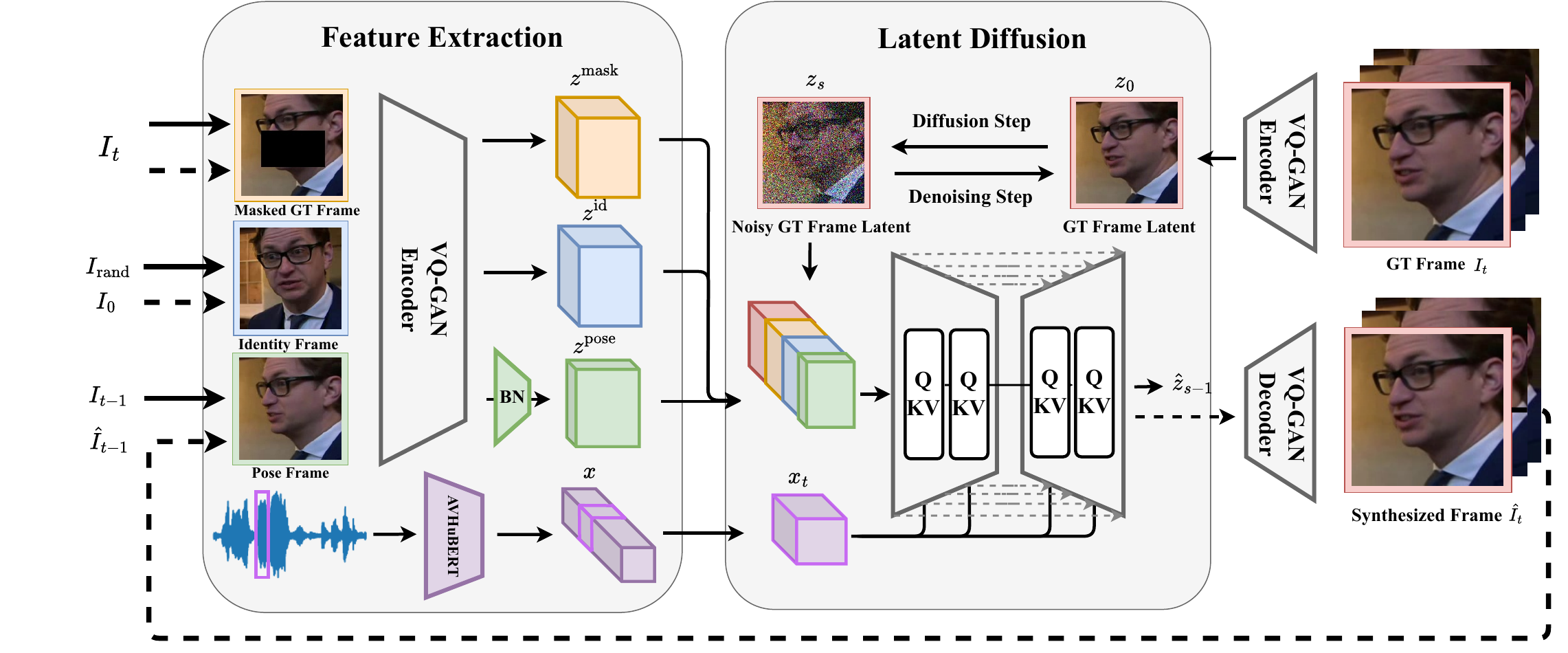}
    \caption{Architecture of the \textsc{SyncDiff} network. Solid lines denote inputs during training while dashed lines denote inputs during inference. $I_t$ denotes the ground-truth frame at timestep t, $\hat{I}_t$ denotes the synthesized frame at timestep t, and $I_{rand}$ denotes randomly sampled frame from the groundtruth sequence. BN denotes the bottleneck layer which compresses the dimension of the pose prior.}
    \label{fig:Overall}
    \vspace{-0.2cm}
\end{figure*}

\subsection{Audio Features for Talking Head Synthesis}
As the audio features serve as the driving factor for video frame synthesis, it is crucial to enhance their quality in encoding temporal and lip movement enabling information. Instead of using the Mel-spectrogram features that have a limited context window as in \textsc{Wav2Lip} and \textsc{SadTalker}~\cite{Zhang2022SadTalkerLR}, \textsc{DiffTalk} chooses the \textsc{DeepSpeech}~\cite{Hannun2014DeepSS} features, which uses bidirectional recurrent layers to encode speech spectrograms. With recent advances in self-supervised speech representations, audio features extracted from pretrained acoustic models such as \textsc{Wav2vec2}~\cite{baevski2020wav2vec}, \textsc{HuBERT}~\cite{hsu2021hubert} have been shown to encode richer contextual information and have exhibited great performance advantages on various downstream tasks over traditional methods~\cite{Yang2021SUPERBSP}. \textsc{DAETalker} uses \textsc{Wav2vec2} features as speech representations for talking head synthesis. These aforementioned audio features are purely audio-driven. \textsc{AVHuBERT} pretrains a self-supervised \textsc{HuBERT}-like Transformer-based model using masked multimodal cluster prediction objective on speech-video datasets to learn speech and video representations that have better correlations between the two modalities. \textsc{TalkLip}~\cite{Wang2023SeeingWY} uses \textsc{AVHuBERT} features to drive the talking head generations and achieve the SOTA synchronization performance.

\subsection{Diffusion Model}

Denoising diffusion probabilistic models (DDPMs) have recently demonstrated great power in image synthesis \cite{ho2020denoising,NEURIPS2021_49ad23d1} and image impainting \cite{lugmayr2022repaint} tasks. 

The main drawback of DDPM models is their high computational cost in both training and inference due to the pixel-based input and output space~\cite{ronneberger2015u} and large number of diffusion steps~\cite{ho2020denoising}. To mitigate the cost incurred by operating on pixel space, latent diffusion models (LDMs)~\cite{rombach2022high} are proposed, which use a VQGAN autoencoder~\cite{esser2021taming} to reduce the high-resolution image to low-dimensional latent space for the training of the DDPMs. LDMs have been applied to many multimodal tasks including text-to-image translation~\cite{rombach2022text,kawar2023imagic}, text-to-audio generation~\cite{huang2023make} and speech-to-video generation~\cite{Shen2023DiffTalkCD}, etc. To reduce computational cost resulting from a large number of diffusion steps, denoising diffusion implicit models (DDIM)~\cite{song2020denoising}, consistency models~\cite{Song2023ConsistencyM} and EDM~\cite{Karras2022ElucidatingTD} are proposed. DDIM generalize the Markovian diffusion process assumed by DDPM into a class of non-Markovian diffusion process that lead to the same training objective as that of DDPM but with shorter Markov chains and thus fewer diffusion steps. DDIM follows the same training process as DDPM, it can be directly applied to accelerate the inference process of DDPM-based models~\cite{Shen2023DiffTalkCD,huang2022prodiff}.

\section{Method}
\label{sec:method}

\subsection{Overview}

A typical diffusion-based talking head synthesis model like \textsc{DiffTalk} is conditioned on a given audio clip and a reference frame where the reference frame is used as a driving factor of the human identity in the generated videos. 
However, its deficiency in audio-visual synchronization is evident, suggested by its low \textsc{SyncNet} scores and high lip-reading word error rate. In this paper, we propose to incorporate a pose frame that serves as temporal visual priors to further improve the synthesized videos' audio-visual synchronization. We additionally utilize a bottleneck layer over the pose frame to prevent the model from learning shortcuts. The overall architecture is presented in Figure~\ref{fig:Overall}.

\subsection{Audio Feature Extraction}
We use \textsc{AVHuBERT} to extract facial-informative audio features, which is a self-supervised multimodal Transformer pretrained on audio-visual datasets. Specifically, we first extract the Mel-spectrogram with a feature rate of $100$ Hz using a moving window that has a window size of $25$ ms and a hop-size of $10$ ms. Every four consecutive Mel-spectrogram features are grouped as one feature to be input to a linear layer, resulting in the grouped Mel-spectrogram features with a rate of $25$ Hz, the same rate as the video frames. The grouped features are then input to $12$ layers of \textsc{AVHuBERT} Transformer layers to extract contextualized audio features. The first $9$ layers of \textsc{AVHuBERT} are frozen during training, and the remaining $3$ layers are jointly trained with the diffusion model. To compare \textsc{AVHuBERT} features with other audio features, we additionally extract \textsc{DeepSpeech} features following the preprocessing in \textsc{DiffTalk}. The extracted features are fed to a learnable temporal filtering network~\cite{thies2020neural} for improved inter-frame consistency.

The audio features corresponding to the target frame are used as the conditioning inputs to the diffusion model.

\subsection{Temporal-Augmented Visual Priors}

Previous works~\cite{KR2020ALS,Wang2023SeeingWY,Shen2023DiffTalkCD,Du2023DAETalkerHF} generally use the masked ground-truth (GT) frame, a randomly sampled reference frame and the audio features as priors to condition the GAN-based or diffusion-based generator during training, where the masked frame provides the head pose information and the reference frame provides the appearance identity of the lip region. During testing, the given first frame~\cite{KR2020ALS,Wang2023SeeingWY} or the synthesized previous frame~\cite{Shen2023DiffTalkCD} replaces the randomly sampled frame as the reference frame to provide identity information.

We argue that the model trained with a randomly sampled reference frame fails to learn the temporal transition between consecutive frames, which leads to inter-frame incoherence. Such incoherence is especially evident in the diffusion-based model as discussed in \textsc{DiffTalk}. \textsc{DiffTalk} mitigate the incoherence by training with random reference but inferring with generated previous frames. However, doing so introduces a train-test distribution mismatch, which can negatively influence the model's synchronization performance.
One naive approach is to use the \textit{previous frames} immediately preceding the target frame, instead of randomly sampled ones, as reference frames to provide the temporal information during training. However, this approach incurs significant model degradation because the neighborhood frames are usually so similar that the model can achieve a low loss by directly copying the reference frame. Such degradation is referred to as training shortcut in previous works~\cite{KR2020ALS,Shen2023DiffTalkCD}. 

We discover that applying a simple bottleneck layer over the temporal pose frame resolves the shortcut issue. The bottleneck layer is a 2D convolutional layer and is trained to compress the reference features so that they contain mainly temporal pose priors. Furthermore, we propose to incorporate two reference frames, in addition to the masked GT frame, as priors to provide identity and temporal pose information separately. During training, one random sampled frame serves as the identity reference frame and one previous frame with bottleneck serves as the temporal pose frame, while during inference, the given first frame and the generated previous frame serve as the identity and pose frame. 

The proposed triple-prior approach has two advantages compared to the traditional double-prior approach. Firstly, the explicit separation of speaker identity and temporal lip pose information into two sources can lead to a  distribution easier to learn. Our empirical results in Sec~\ref{sec:variance} suggest that the triple-prior model can more effectively disentangle identity and temporal pose information while remaining invariant to the change of other aspects, e.g. lighting condition and camera position. Secondly, compared to \textsc{DiffTalk} approach, our triple prior approach has unified reference frame distribution during training and inference.

\subsection{Conditional Latent Diffusion}
We model the generation process as a denoising process using a LDM. To project between the pixel space and the latent space, we adopt the VQGAN-based autoencoder~\cite{esser2021taming,rombach2022high,Shen2023DiffTalkCD}, which consisting of a pair of encoder $\mathcal{E}$ and decoder $\mathcal{D}$ with a downsampling and upsampling factor of $4$, respectively. As discussed in the previous section, three image frames that act as visual priors will go through the encoder $\mathcal{E}$ whose output space is $h \times 
w \times 3$. The temporal pose prior will go through one additional bottleneck convolution layer to further compress its dimension before it's concatenated with the other two priors. The visual feature extraction step can be formulated as:
\begin{align}
&E_{m}, E_{i}, E_{p} = \mathcal{E}(I_{m}, I_{i},I_{p})\\
&E_{v} = \mathrm{Concat}(E_{m}, E_{i}, \mathrm{BN}(E_{p}))
\end{align}
where $I_{m}, I_{i}, I_{p}$ are respectively the masked frame, identity frame, and pose frame and $\mathrm{BN}$ is the bottleneck layer.

The extracted visual and audio priors $E_v$, $E_a$ are then fed into a denoising \textsc{UNet}~\cite{ronneberger2015u} $\mathcal{M}$ together with a sampled timestep d and the respective $z_d$ which denotes the latent representation of the target at the d timestep of the forward diffusion process. The architecture of the entire SyncDiff method is shown in \ref{fig:Overall}. The loss function can then be formulated as
\begin{equation}
    \mathcal{L}_{LDM} := \mathbb{E}_{z, \epsilon \sim \mathcal{N}(0,1),E_v, E_a, d} [\lVert\epsilon - \mathcal{M}(z_d, d, E_v, E_a)\rVert^2]
\end{equation}

Our conditional \textsc{UNet} utilizes a cross-attention mechanism to learn across visual and audio modalities. The visual prior $E_v \in \mathcal{R}^{h \times w \times 7}$ is further concatenated with the noisy latent $z_d$ and form an embedding $E_q \in \mathcal{R}^{h \times w \times 10}$ which is directly fed into the network and used to calculate the query of the cross-attention. The audio feature $E_a$ is fed into the intermediate layers and used to calculate the key and value. 

Denote the estimation of the latent at timestep d as $\tilde{z}_d$. During inference, the denoised latent at the final step $d = 0$ can be decoded back to the pixel space using the decoder $\mathcal{D}$.
\begin{equation}
    I_{pred} := \mathcal{D}(\tilde{z}_0)
\end{equation}

\begin{table*}[ht!]
  \centering
  \begin{sc}
  \setlength{\tabcolsep}{1.6pt}
  \begin{tabular}{l|cccccc|cccccc}
    \toprule
     & \multicolumn{6}{c|}{LRS2} &\multicolumn{6}{c}{LRS3} \\
    \cmidrule(lr){2-7}\cmidrule(lr){8-13} 
    Method & PSNR$\uparrow$ & SSIM$\uparrow$ & LPIPS$\downarrow$ & LSE-C$\uparrow$ & LSE-D$\downarrow$ & WER$\downarrow$ & PSNR$\uparrow$ & SSIM$\uparrow$ & LPIPS$\downarrow$ & LSE-C$\uparrow$ & LSE-D$\downarrow$ & WER$\downarrow$\\\midrule\midrule
    GT & N/A & 1.000 & 0.000 & 8.25 & 6.26 & 23.82 & N/A & 1.000 & 0.000 & 7.63 & 6.89 & 41.04\\
    GT-Rec & 35.30 & 0.956 & 0.022  & 8.02 & 6.35 & 28.47 & 36.06 & 0.958 & 0.015 & 7.48 & 6.95 & 44.44 \\\midrule
    Wav2Lip & 31.07 & 0.859 & 0.079 & \second{7.83} & 6.65 & 90.67 & 31.04 & \second{0.846} & 0.079 & \second{8.10} & \second{6.68} & 91.58 \\
    SadTalker & 29.27 & 0.543 & 0.109 & 5.45 & 8.34 & 110.64 & 29.08 & 0.641 & 0.144 & 5.27 & 8.65 & 107.13 \\
    TalkLip & 31.16 & 0.850 & 0.084 & \first{8.53} & \first{5.70} & \first{23.43} & 31.10 & \first{0.852} & 0.084 & \first{8.11} & \first{6.41} & \first{22.01} \\
    DiffTalk & \second{32.36} & \second{0.873} & \second{0.056} & 6.11 & 7.93 & 114.34 & \second{31.12} & 0.779 & \second{0.076} & 4.80 & 9.20 & 114.16 \\
    \algname & \first{32.39} & \first{0.874} & \first{0.049} & 7.80 & \second{6.58} & \second{67.40} & \first{32.18} & \second{0.846} & \first{0.058} & 6.82 & 7.58 & \second{83.94}\\
    \bottomrule
  \end{tabular}
  \vspace{-0.2cm}
  \caption{Comparison of visual quality and lip synchronization scores on  LRS2 and LRS3 datasets. \textsc{GT-Rec} are ground-truth videos encoded and reconstructed using our image autoencoder and serve as a topline for our \algname models. The best results are highlighted in bold \first{black} and the second best results are highlighted in bold \second{violet}.}
  \label{tab:main_results}
  \end{sc}
\end{table*}

\section{Experiments}
\label{sec:experiments}

\subsection{Datasets}
We conduct our experiments using two canonical datasets, LRS2~\cite{afouras2018deep} for training and evaluation and LRS3~\cite{Afouras2018LRS3TEDAL} for evaluation only. LRS2 dataset contains around $29$ hours of talking-head videos with the ground-truth transcripts of the speakers' speech. We follow the standard train/valid/test split in LRS2 dataset. LRS3 dataset contains thousands of spoken sentences from TED and TEDx videos. We use only the test set for evaluation, which contains $1321$ video samples with a total duration of $51$ minutes.

\subsection{Data Preprocessing}
Videos in both datasets have a frame rate of $25$ frames per second (fps) and $3$ RGB channels. The LRS2 videos have a dimension of $160\times160\times3$ and LRS3 videos have a dimension of $224\times224\times3$. During training, all videos are resized to a resolution of $256\times256\times3$.
Following \textsc{Wav2Lip}~\cite{KR2020ALS} and \textsc{TalkLip}~\cite{Wang2023SeeingWY}, we use \textsc{Face-Alignment}~\cite{bulat2017far} to detect face region and use the lower half of the face region as lip region.

Audios are resampled to $16000$ Hz. We extract Mel-spectrogram features with $10$-ms hop sizes of $25$-ms window lengths and $26$-filter filter bank, resulting in spectrogram features with $26$ bins and a feature rate of $100$ Hz. We then group four consecutive features together to create $25$-Hz spectrogram features with a dimensionality of $104$. To replicate \textsc{DiffTalk}~\cite{Shen2023DiffTalkCD} results, we additionally extract $25$-Hz \textsc{DeepSpeech}~\cite{Hannun2014DeepSS} features with dimensions of $16\times29$ following the practice in \textsc{VOCA}~\cite{Cudeiro2019CaptureLA}. We apply a window of $16$ consecutive features centered at each timestamp to create $25$-Hz \textsc{DeepSpeech} features with a dimensionality of $16\times16\times29$.

\subsection{Implementation Details}
We use a VQGAN-based image autoencoder~\cite{rombach2022high} with a downsampling rate of $4$ to compress the input images from the resized dimension of $256\times256\times3$ to a latent dimension of $64\times64\times3$, which is the input dimension of the LDM. We apply a $1\times1$ 2D convolution layer on the  channel dimension of the latent embedding of the previous frame to further compress its dimension to $64\times64\times1$. The conditioning input of the LDM has the same dimension as the audio features. In the case of \textsc{AVHuBERT} features, the audio features have a dimension of $768$. In the case of \textsc{DeepSpeech} features, the audio features have a dimension of $128$, which is the same dimension as in \textsc{DiffTalk}. The number of diffusion steps of LDM is set to $1000$ during training and the number of steps of the DDIM sampler is set to $200$ during inference. The entire \algname model is trained for $200$ epochs on $6$ GPUs with a batch size of $8$ on each GPU.

\subsection{Metrics}
Following recent works on talking head generation ~\cite{KR2020ALS,park2022synctalkface,Du2023DAETalkerHF,Wang2023LipFormerHA,Wang2023SeeingWY,Shen2023DiffTalkCD}, we use peak signal-to-noise ratio (PSNR), structural similarity index measure (SSIM) and learned perceptual image patch similarity (LPIPS)~\cite{zhang2018perceptual} scores to measure visual quality, use ``lip sync error - confidence" (LSE-C) and ``lip sync error - distance" (LSE-D) scores generated by a pretrained \textsc{SyncNet}~\cite{Chung16a} to measure lip-speech synchronization and use word error rate (WER) to measure lip reading intelligibility. We use \textsc{AVHubert-large}, a lip-reading model finetuned on LRS2 dataset, to transcribe the generated video, and calculate WER based on the resulting transcripts. 

\subsection{Comparison Methods} 
We compare \algname against four SOTA methods: \textsc{Wav2Lip}~\cite{KR2020ALS}, \textsc{SadTalker}~\cite{Zhang2022SadTalkerLR}, \textsc{TalkLip}~\cite{Wang2023SeeingWY} and \textsc{DiffTalk}~\cite{Shen2023DiffTalkCD}. \textsc{Wav2Lip} is a GAN-based model with its discriminator being a pretrained \textsc{SyncNet}-based lip sync model and achieves high lip synchronization scores. \textsc{SadTalker} is a 3D-based talking face model conditioned on generated realistic 3D pose coefficients. \textsc{TalkLip} improves on \textsc{Wav2Lip} by replacing Mel-spectrogram features with pretrained \textsc{AVHuBERT} features and replacing \textsc{SyncNet} lip-syncing model with \textsc{AVHuBERT} lip-reading model. \textsc{DiffTalk} uses diffusion models to generate face images and achieves the SOTA visual quality on talking head generation.

\section{Results}
\subsection{Main Results}
Table~\ref{tab:main_results} compares \algname against ground truth videos, ground truth videos reconstructed using the image autoencoder, and four state-of-the-art models quantitatively. The performances are measured using visual quality (PSNR, SSIM and LPIPS), lip-speech synchronization (LSE-C and LSE-D), and lip reading intelligibility (WER) scores. on the in-domain LRS2 dataset, we make three key observations. First, our \algname can generate talking head videos with the best image quality compared to GAN-based \textsc{Wav2Lip} and \textsc{TalkLip} and diffusion-based \textsc{DiffTalk}. Second, \algname, trained with \textsc{AVHuBERT} features and additional previous frame, achieves significantly better synchronization scores than \textsc{DiffTalk}, and the LSE-C and LSD-D scores of \algname approach to the reconstructed ground truth topline. Finally, if we exclude \textsc{TalkLip}, which explicitly trains using paired video and text data, \algname outperforms other methods that do not use additional text information on lip-reading intelligibility.

The same observation generally holds on the out-of-domain LRS3 dataset. However, we do notice a more evident drop in synchronization using diffusion-based \textsc{DiffTalk} and \algname compared to GAN-based \textsc{Wav2Lip} and \textsc{TalkLip}, which, we hypothesize, is due to \algname reconstruct lip movement as well as the surrounding texture while \textsc{Wav2Lip} and \textsc{TalkLip} focus less on texture but more on lip movement. Such behavior may limit the generalizability of \algname to unseen identities.

We should make a side note that we use the officially released \textsc{SadTalker} checkpoint, which is not pretrained on LRS2. This choice possibly leads to its suboptimal performance in our evaluation.

\subsection{Perceptual Evaluation}

\begin{figure*}[ht!]
    \centering
    \begin{subfigure}[b]{0.5\textwidth}
         \centering
         \includegraphics[width=\textwidth,trim={0 0 0 0cm},clip]{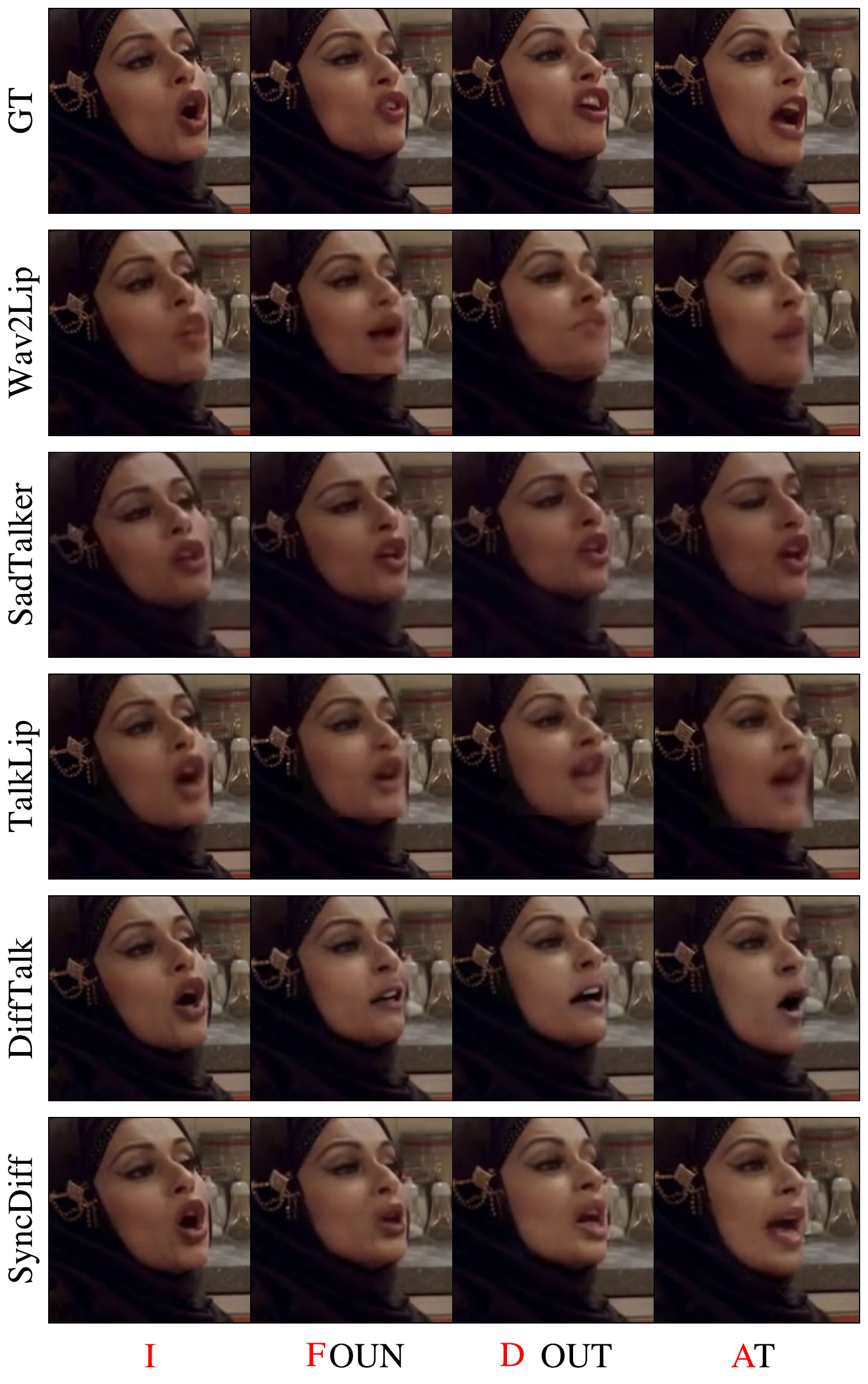}
     \end{subfigure}
     \hfill
     \begin{subfigure}[b]{0.478\textwidth}
         \centering
         \includegraphics[width=\textwidth,trim={1.2cm 0 0 0},clip]{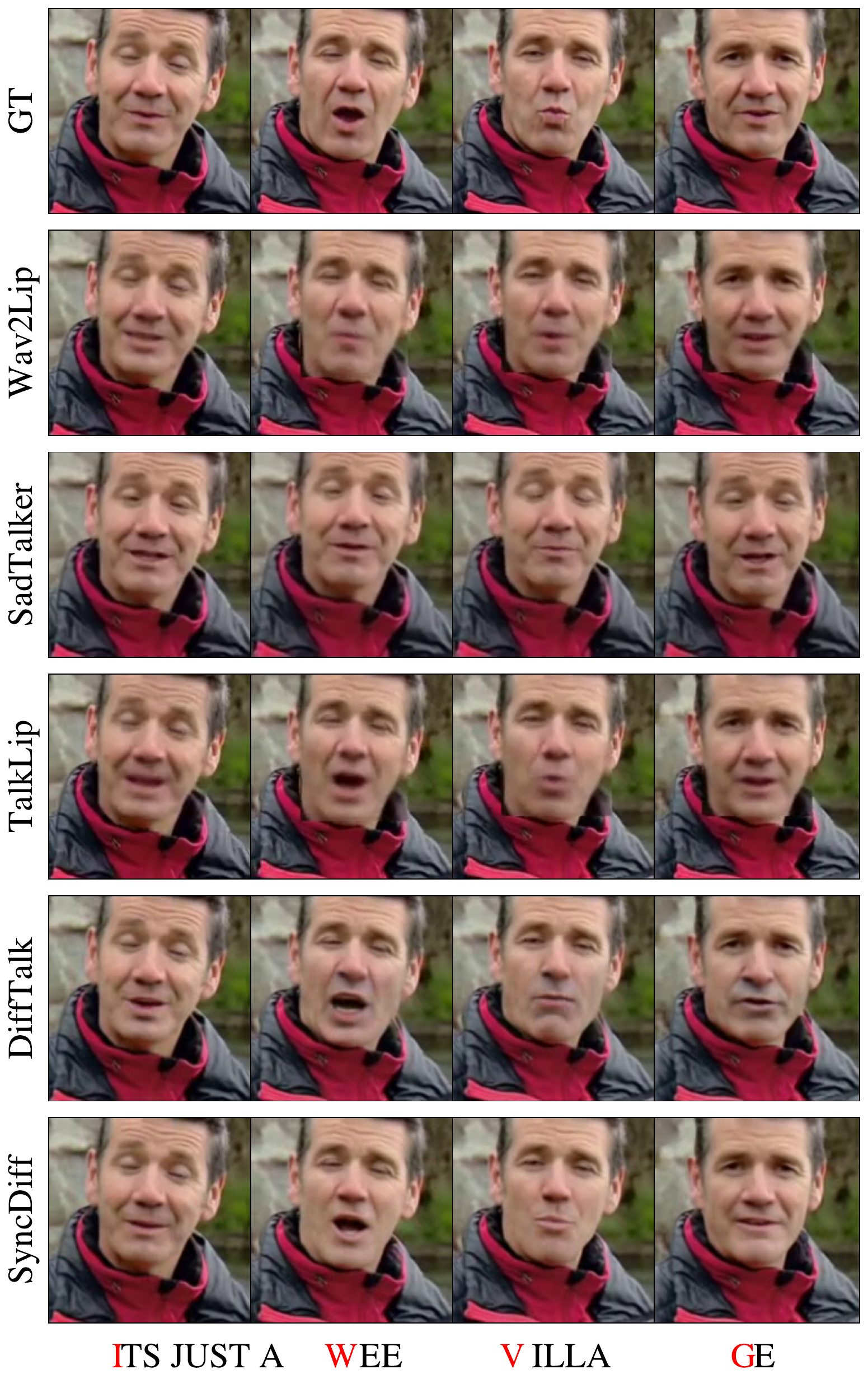}
     \end{subfigure}
    \vspace{-0.2cm}
    \caption{Visual comparison with SOTA talking head generation methods. The letter that each frame corresponds to is marked in {\color{red}{red}}.}
    \label{fig:qualitative}
    \vspace{-0.3cm}
\end{figure*}

We provide uniformly sampled snapshots of two generated talking face videos in Figure~\ref{fig:qualitative} to give a better visual demonstration of the advantage of \algname compared against other SOTA methods. We can see that compared to GAN-based \textsc{Wav2Lip} and \textsc{TalkLip}, Diffusion-based \textsc{DiffTalk} and \algname generate the talking face frames with sharper textures and more details. Also, because the \textsc{DiffTalk} and \algname are trained to generate the entire image instead of the face region, we do not see an obvious boundary mismatch at the face region that appears frequently in videos generated by \textsc{Wav2Lip} and \textsc{TalkLip}. Comparing \textsc{DiffTalk} and \algname, we observe that video frames from the latter exhibit lip shapes closer to the ground truth frames than those from the former, for example, at frames for ``A'', ``V'', ``G''.

\begin{figure}[t!]
    \centering
    \includegraphics[width=0.47\textwidth,trim={0 0cm 0 0},clip]{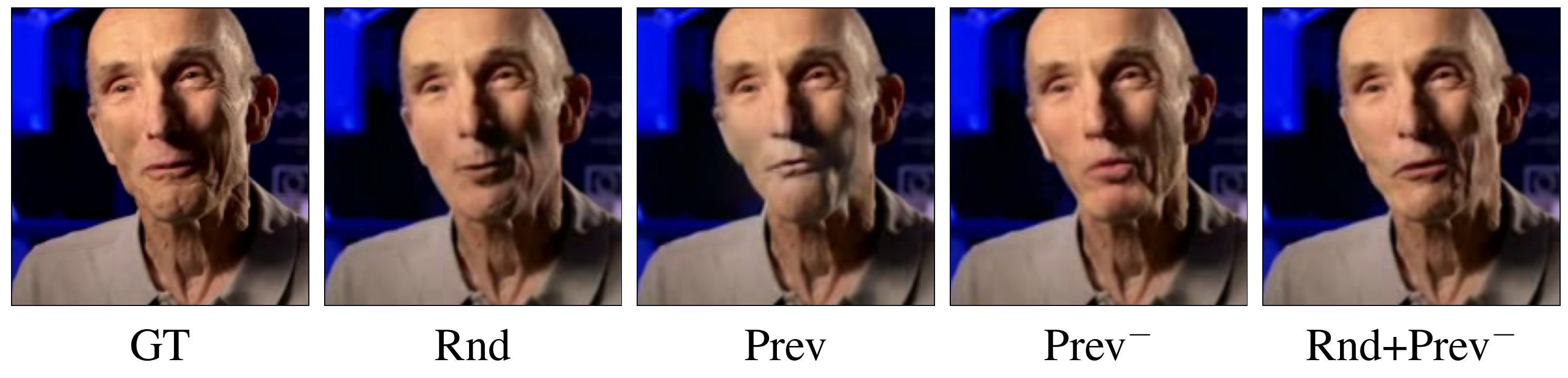}
    \vspace{-0.2cm}
    \caption{Visual comparison of using different reference frames.}
    \label{fig:ref_frame}
\end{figure}

\subsection{Comparison on Audio Features}
We compare the performance of using \textsc{DeepSpeech} and \textsc{AVHuBERT} audio features. The results are shown in Table~\ref{tab:audio_feature}. We observe that the synchronization and intelligibility scores are greatly improved using \textsc{AVHuBERT} features. We believe this is because RNN-based \textsc{DeepSpeech} audio representations have three drawbacks compared to Transformer-based \textsc{AVHuBERT}. First, \textsc{DeepSpeech} contains fewer parameters than \textsc{AVHuBERT} and therefore lower model capacities. Second, \textsc{DeepSpeech} uses bidirectional RNN layers, that do not sufficiently model long-term context compared to self-attention layers, resulting in better audio representations and better overall video quality. Finally, \textsc{DeepSpeech} is trained on pure audio while \textsc{AVHuBERT} is trained to match the audio representation with corresponding lip shapes. With additional facial priors, the audio features extracted from the latter are more conducive to lip movement generation and thus higher synchronization and intelligibility scores. 

\begin{table}
  \centering
  \begin{sc}
  \setlength{\tabcolsep}{1.5pt}
  \begin{tabular}{l|cccccc}
    \toprule
    Feat & PSNR$\scriptstyle\uparrow$ & SSIM$\scriptstyle\uparrow$ & LPIPS$\scriptstyle\downarrow$ & LSE-C$\scriptstyle\uparrow$ & LSE-D$\scriptstyle\downarrow$ & WER$\scriptstyle\downarrow$\\\midrule\midrule
    DS & 32.36 & 0.873 & {0.056} & 6.11 & 7.93 & 114.34 \\
    AVH & \textbf{32.40} & \textbf{0.877} & \textbf{0.055} & \textbf{7.05} & \textbf{7.11} & \textbf{90.24} \\
    \bottomrule

  \end{tabular}
  \caption{Comparison between \textsc{DeepSpeech} (DS) and \textsc{AVHuBERT} (AVH) speech features with randomly sampled video frame as visual prior during training.}
  \label{tab:audio_feature}
  \end{sc}
\end{table}

\subsection{Effect of Reference Frames}\label{sec:variance}

\begin{table}
  \centering
  \begin{sc}
  \begin{small}
  \setlength{\tabcolsep}{1.3pt}
  \begin{tabular}{l|cccccc}
    \toprule
    Ref Frame & PSNR$\scriptstyle\uparrow$ & SSIM$\scriptstyle\uparrow$ & LPIPS$\scriptstyle\downarrow$ & LSE-C$\scriptstyle\uparrow$ & LSE-D$\scriptstyle\downarrow$ & WER$\scriptstyle\downarrow$\\\midrule\midrule
    Rnd & \textbf{32.40} & \textbf{0.877} & 0.055 & 7.05 & 7.11 & 90.24 \\
    Prv & 31.61 & 0.851 & 0.057 & 7.38 & 6.91 & 76.16 \\
    Prv$^-$ & 32.21 & 0.866 & \second{0.053} & \textbf{7.81} & \textbf{6.55} & \textbf{62.33} \\
    Rnd+Prv$^-$ & \second{32.39} & \second{0.874} & \textbf{0.049} & \second{7.80} & \second{6.58} & \second{67.40} \\
    \bottomrule
  \end{tabular}
  \end{small}
  \caption{Comparison between using different video frames as input to predict the next frame during training. \textsc{Rnd}, \textsc{Prv} and \textsc{Prv}$^-$ denotes using a randomly selected frame, previous frame and the bottlenecked previous frame, respectively. The best results are highlighted in bold \first{black} and the second best results are highlighted in bold \second{violet}.}
  \label{tab:ref_frame}
  \end{sc}
\end{table}
We conduct experiments to compare the effect of using different reference frames to condition the diffusion generator: a randomly sampled video frame (\textsc{Rnd}), the previous frame of the target frame (\textsc{Prev}), the bottlenecked previous frame (\textsc{Prev}$^-$), and the combination of the random and bottlenecked previous frame (\textsc{Rnd}+\textsc{Prev}$^-$). The objective results are shown in Table~\ref{tab:ref_frame} and the snapshots of the generated videos are shown in Figure~\ref{fig:ref_frame}. By comparing the \textsc{Rnd} and the \textsc{Prev} results, we verify the existence of the short-cut issue that leads to the grey-out of the lip region. By comparing \textsc{Prev} and \textsc{Prev}$^-$, we observe that the bottleneck layer after the previous frame resolves this issue. By comparing \textsc{Rnd} and \textsc{Prev}$^-$, we observe that using the bottlenecked previous frame greatly improves lip synchronization with a slight sacrifice of image quality. The best strategy is \textsc{Rnd}+\textsc{Prev}$^-$, which combines the advantages of the two to achieve both high image quality and high synchronization.

To further study the behavior of \textsc{SyncDiff} with the additional input of the previous frame, we measure the variance of latent representations encoded by the \textsc{UNet} of the diffusion model, across different identity/pose frames that share the same identity/pose. Specifically, Two sets of experiments, namely \textsc{Ident} and \textsc{Pose}, are conducted on two models, \textsc{Rnd} and \textsc{Rnd+Prev}$^-$. In \textsc{Ident}, we randomly select $10$ videos from LRS2. For each video, we choose the second frame to be the target and fix the pose frame to be the first frame. The subsequent frames in the same video serve as the identity frames sharing the same speaker identity. One latent representation per identity frame is extracted from the diffusion model. We compute the variances in each dimension and average them across dimensions and videos. In \textsc{Pose}, we choose $5$ videos from LRS2 featuring the same speaker uttering the same sentence: ``Thanks for watching''. Frames corresponding to the phone ``wa'' are manually labeled and treated as pose frames sharing the same pose. We set the identity frames as the first frame of the first video, select the frame after each pose frame as the GT frame, and vary the masked GT frame and the pose frame across $5$ videos. Similar to \textsc{Ident} experiment, we compute the averaged variance across latent dimensions. The results are shown in Table~\ref{tab:variance}. In both experiments, we observe \textsc{Rnd+Prev}$^-$ has notably lower variance. This observation implies that the latent features with \textsc{Rnd+Prev}$^-$ more effectively capture both identity and pose-related information, remaining invariant to changes in other aspects.

\begin{table}
  \centering
  \begin{sc}
  \begin{tabular}{l|cc|cc}
    \toprule
    & \multicolumn{2}{c|}{Ident} & \multicolumn{2}{c}{Pose} \\
    \cmidrule(lr){2-3}\cmidrule(lr){4-5} 
    Ref & Rnd & Rnd+Prv$^-$ & Rnd & Rnd+Prv$^-$\\\midrule\midrule
    Var$\scriptstyle\downarrow$ & 0.340 & \textbf{0.075} & 0.530 & \textbf{0.328} \\
    \bottomrule
  \end{tabular}
  \caption{Comparison between the average variance over latent dimensions when varying identity frame or pose frame while keeping the other fixed. \textsc{Ident} denotes varying the identity frames and fixing the pose frames and \textsc{Pose} denotes the reverse. \textsc{Var} denotes the averaged variance.}
  \label{tab:variance}
  \end{sc}
\end{table}

\subsection{Ablation Study on Number of Previous Frames}
We experiment on providing more temporal information to \algname by increasing the number of previous frames fed to the LDM. The results are shown in Table~\ref{tab:nprev}. Unfortunately, we observe no improvement in lip-synchronization; Using one previous frame as a reference yields the best performance. We believe this is because the LDM architecture lacks recurrent or self-attention layers over the input sequence to capture temporal dependencies of lip movement, which we leave as future works to explore.
\begin{table}
  \centering
  \begin{sc}
  \setlength{\tabcolsep}{2.3pt}
  \begin{tabular}{l|cccccc}
    \toprule
    NP & PSNR$\scriptstyle\uparrow$ & SSIM$\scriptstyle\uparrow$ & LPIPS$\scriptstyle\downarrow$ & LSE-C$\scriptstyle\uparrow$ & LSE-D$\scriptstyle\downarrow$ & WER$\scriptstyle\downarrow$\\\midrule\midrule
    1& 32.39 & \first{0.874} & \first{0.049} & \first{7.80} & \first{6.58} & \first{67.40} \\
    5 & 32.34 & 0.869 & 0.052 & 7.46 & 6.85 & 84.13 \\
    10 & \first{32.45} & {0.872} & 0.051 & 7.48 & 6.82 & 86.20 \\
    \bottomrule
  \end{tabular}
  \caption{Comparison on using different number previous frames (NP) as input to LDM.}
  \label{tab:nprev}
  \end{sc}
\end{table}

\begin{table}[h]
  \caption{Comparison on freezing different number of layers (NF) in \textsc{AVHuBERT}.}
  \label{tab:freeze}
  \centering
  \begin{sc}
  \setlength{\tabcolsep}{2.3pt}
  \begin{tabular}{l|cccccc}
    \toprule
    NF & PSNR$\scriptstyle\uparrow$ & SSIM$\scriptstyle\uparrow$ & LPIPS$\scriptstyle\downarrow$ & LSE-C$\scriptstyle\uparrow$ & LSE-D$\scriptstyle\downarrow$ & WER$\scriptstyle\downarrow$\\\midrule\midrule
    3 & 32.35 & 0.872 & 0.050 & 7.56 & 6.75 & 80.90 \\
    6 & 32.37 & \first{0.874} & \first{0.049} & 7.64 & 6.67 & 72.33 \\
    9 & \first{32.39} & \first{0.874} & \first{0.049} & \first{7.80} & \first{6.58} & \first{67.40} \\
    12 & 31.97 & 0.855 & 0.057 & 7.46 & 6.87 & 75.92 \\
    \bottomrule
  \end{tabular}
  \end{sc}
  \vspace{-0.3cm}
\end{table}

\subsection{Ablation Study on Freezing AVHuBERT}
\textsc{AVHuBERT}
It is known that the features from different layers of \textsc{HuBERT} based model capture different levels of linguistic information~\cite{shi2022avhubert}.
We therefore experiment on freezing the first few layers of the pretrained $12$-layer \textsc{AVHuBERT} model during the training. The results are shown in Table~\ref{tab:freeze}. We observe a U-shape curve in performance with the increase of number of frozen layers and freezing the first nine layers of \textsc{AVHuBERT} yields the best overall performance.

\section{Limitation}
Our model mainly suffers from three limitations. Firstly, the inference of the diffusion model is slow. Although we apply DDIM sampler to speed up the inference, it still takes a significant amount of time to generate one single video compared to GAN-based methods. Diffusion models with fewer diffusion steps such as consistency models and EDM can be explored to further speed up the inference.
Secondly, although \algname achieves significant improvement in lip synchronization, we observe an evident performance degradation on out-of-distribution testing data, compared to the GAN-based methods. Adding explicit cross-modality contrastive loss for better audio-visual representation as in \textsc{Wav2Lip} and \textsc{TalkLip} might be one promising solution. However, it is not easy to directly apply such loss in diffusion models because of the heavy noise in the intermediate diffusion steps. We leave this as future works to explore. 
Last but not least, our ablation studies show that one previous frame is a better temporal prior than $5$ or $10$ previous frames, suggesting \algname does not leverage long-term temporal patterns efficiently. Self-attention or recurrent layers can be experimented to extract better temporal features.
\label{sec:limitation}

\section{Conclusion}
\label{sec:conclusion}
In this work, we present \textsc{SyncDiff}, a diffusion-based talking head synthesis model that simultaneously achieves both high visual quality and good synchronization. Our contribution can be summarized in mainly three aspects. 
First, we introduce a bottleneck layer to incorporate visual temporal pose information in the diffusion-based talking head synthesis model, which resolves the learning short-cut issue of using previous frames during training and leads to a substantial improvement in lip synchronization. This modification opens the possibility of incorporating further temporal video information in the training of existing GAN-based and diffusion-based talking head generators.
Second, we leverage a self-supervised audio-visual pretrained model, \textsc{AVHuBERT}, to facilitate diffusion-based talking head generation which contributes to further enhanced lip synchronization. Finally, We conduct extensive experiments to demonstrate the advantage of \textsc{SyncDiff} over other state-of-the-art methods in terms of visual quality and lip synchronization.

{\small
\bibliographystyle{ieee_fullname}
\bibliography{main}
}

\end{document}